# Configurable 3D Scene Synthesis and 2D Image Rendering with Per-Pixel Ground Truth using Stochastic Grammars

**Chenfanfu Jiang*** · **Siyuan Qi*** · **Yixin Zhu*** · **Siyuan Huang*** · **Jenny Lin** · **Lap-Fai Yu** · **Demetri Terzopoulos** · **Song-Chun Zhu**

**Abstract** We propose a systematic learning-based approach to the generation of massive quantities of synthetic 3D scenes and arbitrary numbers of photorealistic 2D images thereof, with associated ground truth information, for the purposes of training, benchmarking, and diagnosing learning-based computer vision and robotics algorithms. In particular, we devise a learning-based pipeline of algorithms capable of automatically generating and rendering a potentially infinite variety of indoor scenes by using a stochastic grammar, represented as an attributed Spatial And-Or Graph, in conjunction with state-of-the-art physics-based rendering. Our pipeline is capable of synthesizing scene layouts with high diversity, and it is configurable inasmuch as it enables the precise customization and control of important attributes of the generated scenes. It renders photorealistic RGB images of the generated scenes while automatically synthesizing detailed, per-pixel ground truth data, including visible surface depth and normal, object identity, and material information (detailed to object parts), as well as environments (*e.g.*, illuminations and camera viewpoints). We demonstrate the value of our synthesized dataset, by improving performance in certain machine-learning-based scene understanding tasks—depth and surface normal prediction, semantic segmentation, reconstruction, *etc.*—and by providing benchmarks for and diagnostics of trained models by modifying object attributes and scene properties in a controllable manner.

* C. Jiang, Y. Zhu, S. Qi, and S. Huang contributed equally to this work. Support for the research reported herein was provided by DARPA XAI grant N66001-17-2-4029, ONR MURI grant N00014-16-1-2007, and DoD CDMRP AMRAA grant W81XWH-15-1-0147.

C. Jiang
SIG Center for Computer Graphics
University of Pennsylvania
E-mail: cffjiang@seas.upenn.edu

S. Qi, Y. Zhu, S. Huang, J. Lin and S.-C. Zhu
UCLA Center for Vision, Cognition, Learning and Autonomy
University of California, Los Angeles
E-mail: {yixin.zhu, syqi, huangsiyuan, jh.lin}@ucla.edu, sczhu@stat.ucla.edu

L.-F. Yu
Graphics and Virtual Environments Laboratory
University of Massachusetts Boston
E-mail: craigyu@cs.umb.edu

D. Terzopoulos
UCLA Computer Graphics & Vision Laboratory
University of California, Los Angeles
E-mail: dt@cs.ucla.edu

## 1 Introduction

Recent advances in visual recognition and classification through machine-learning-based computer vision algorithms have produced results comparable to or in some cases exceeding human performance (*e.g.*, [30, 50]) by leveraging large-scale, ground-truth-labeled RGB datasets [25, 70]. However, indoor scene understanding remains a largely unsolved challenge due in part to the current limitations of RGB-D datasets available for training purposes. To date, the most commonly used RGB-D dataset for scene understanding is the NYU-Depth V2 dataset [112], which comprises only 464 scenes with only 1449 labeled RGB-D pairs, while the remaining 407,024 pairs are unlabeled. This is insufficient for the supervised training of modern vision algorithms, especially those based on deep learning. Furthermore, the manual labeling of per-pixel ground truth information, including the (crowd-sourced) labeling of RGB-D images captured by Kinect-like sensors, is tedious and error-prone, limiting both its quantity and accuracy.

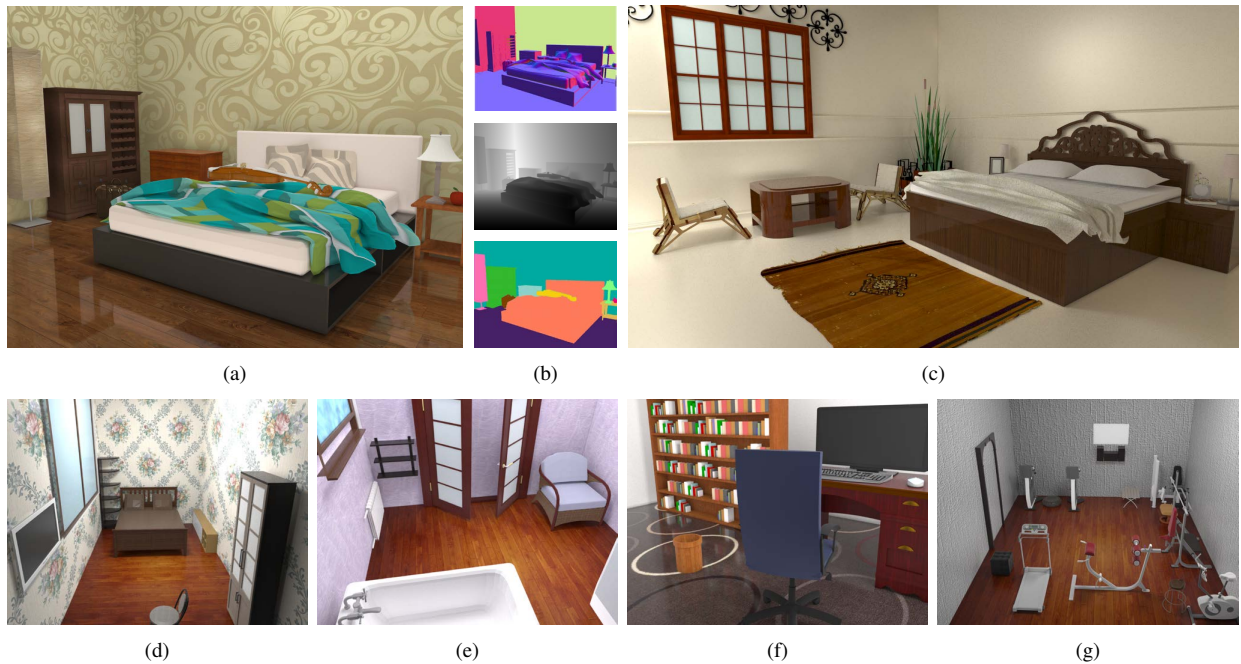

Fig. 1: (a) An example automatically-generated 3D bedroom scene, rendered as a photorealistic RGB image, along with its (b) per-pixel ground truth (from top) surface normal, depth, and object identity images. (c) Another synthesized bedroom scene. Synthesized scenes include fine details—objects (*e.g.*, duvet and pillows on beds) and their textures are changeable, by sampling the physical parameters of materials (reflectance, roughness, glossiness, *etc.*.), and illumination parameters are sampled from continuous spaces of possible positions, intensities, and colors. (d)–(g) Rendered images of four other example synthetic indoor scenes—(d) bedroom, (e) bathroom, (f) study, (g) gym.

To address this deficiency, the use of synthetic image datasets as training data has increased. However, insufficient effort has been devoted to the learning-based systematic generation of massive quantities of sufficiently complex synthetic indoor scenes for the purposes of training scene understanding algorithms. This is partially due to the difficulties of devising sampling processes capable of generating diverse scene configurations, and the intensive computational costs of photorealistically rendering large-scale scenes. Aside from a few efforts in generating small-scale synthetic scenes, which we will review in Section 1.1, a noteworthy effort was recently reported by Song *et al.* [114], in which a large scene layout dataset was downloaded from the Planner5D website.

By comparison, our work is unique in that we devise a complete learning-based pipeline for synthesizing large scale *learning-based configurable* scene layouts via stochastic sampling, in conjunction with photorealistic physics-based rendering of these scenes with associated per-pixel ground truth to serve as training data. Our pipeline has the following characteristics:

- By utilizing a stochastic grammar model, one represented by an attributed Spatial And-Or Graph (S-AOG),

our sampling algorithm combines hierarchical compositions and contextual constraints to enable the systematic generation of 3D scenes with high variability, not only at the scene level (*e.g.*, control of size of the room and the number of objects within), but also at the object level (*e.g.*, control of the material properties of individual object parts).

- As Figure 1 shows, we employ state-of-the-art physics-based rendering, yielding photorealistic synthetic images. Our advanced rendering enables the systematic sampling of an infinite variety of environmental conditions and attributes, including illumination conditions (positions, intensities, colors, *etc.*, of the light sources), camera parameters (Kinect, fisheye, panorama, camera models and depth of field, *etc.*), and object properties (color, texture, reflectance, roughness, glossiness, *etc.*).

Since our synthetic data are generated in a forward manner—by rendering 2D images from 3D scenes containing detailed geometric object models—ground truth information is naturally available without the need for any manual labeling. Hence, not only are our rendered images highly realistic, but they are also accompanied by perfect, per-pixel ground truth color, depth, surface normals, and object labels.

In our experimental study, we demonstrate the usefulness of our dataset by improving the performance of learning-based methods in certain scene understanding tasks; specifically, the prediction of depth and surface normals from monocular RGB images. Furthermore, by modifying object attributes and scene properties in a controllable manner, we provide benchmarks for and diagnostics of trained models for common scene understanding tasks; *e.g.*, depth and surface normal prediction, semantic segmentation, reconstruction, *etc*.

## 1.1 Related Work

*Synthetic image datasets* have recently been a source of training data for object detection and correspondence matching [26, 32, 37, 83, 91, 95, 113, 117, 120, 150], single-view reconstruction [58], view-point estimation [84, 119], 2D human pose estimation [93, 96, 105], 3D human pose estimation [18, 27, 39, 104, 109, 111, 126, 139, 151], depth prediction [118], pedestrian detection [49, 81, 94, 127], action recognition [100, 101, 115], semantic segmentation [103], scene understanding [45, 46, 60, 97], as well as in benchmark datasets [47]. Previously, synthetic imagery, generated on the fly, online, had been used in visual surveillance [98] and active vision / sensorimotor control [122]. Although prior work demonstrates the potential of synthetic imagery to advance computer vision research, to our knowledge no large synthetic RGB-D dataset of *learning-based configurable* indoor scenes has previously been released.

*3D layout synthesis* algorithms [46, 143] have been developed to optimize furniture arrangements based on predefined constraints, where the number and categories of objects are pre-specified and remain the same. By contrast, we sample individual objects and create entire indoor scenes from scratch. Some work has studied fine-grained object arrangement to address specific problems; *e.g.*, utilizing user-provided examples to arrange small objects [33, 144], and optimizing the number of objects in scenes using LARJ-MCMC [140]. To enhance realism, Merrell *et al.* [82] developed an interactive system that provides suggestions according to interior design guidelines.

*Image synthesis* has been attempted using various deep neural network architectures, including recurrent neural networks (RNNs) [41], generative adversarial networks (GANs) [99, 129], inverse graphics networks [65], and generative convolutional networks [77, 136, 137]. However, images of indoor scenes synthesized by these models often suffer from glaring artifacts, such as blurred patches. More recently, some applications of general purpose inverse graphics solutions using probabilistic programming languages have been reported [64, 75, 79]. However, the problem space is enormous, and the quality and speed of inverse graphics "renderings" is disappointingly low and slow.

*Stochastic scene grammar models* have been used in computer vision to recover 3D structures from single-view images for both indoor [72, 147] and outdoor [72] scene parsing. In the present paper, instead of solving visual inverse problems, we sample from the grammar model to synthesize, in a forward manner, large varieties of 3D indoor scenes.

*Domain adaptation* is not directly involved in our work, but it can play an important role in learning from synthetic data, as the goal of using synthetic data is to transfer the learned knowledge and apply it to real-world scenarios. A review of existing work in this area is beyond the scope of this paper; we refer the reader to a recent survey [21]. Traditionally, domain adaptation techniques can be divided into four categories: (i) covariate shift with shared support [11, 20, 42, 51], (ii) learning shared representations [10, 12, 80], (iii) feature-based learning [22, 31, 131], and (iv) parameter-based learning [16, 23, 138, 141]. Given the recent popularity of deep learning, researchers have started to apply deep features to domain adaptation (*e.g.*, [38, 124]).

## 1.2 Contributions

The present paper makes five major contributions:

1. To our knowledge, ours is the first work that, for the purposes of indoor scene understanding, introduces a *learning-based configurable* pipeline for generating massive quantities of photorealistic images of indoor scenes with perfect per-pixel ground truth, including color, surface depth, surface normal, and object identity. The parameters and constraints are automatically learned from the SUNCG [114] and ShapeNet [14] datasets.
2. For scene generation, we propose the use of a stochastic grammar model in the form of an attributed Spatial And-Or Graph (S-AOG). Our model supports the arbitrary addition and deletion of objects and modification of their categories, yielding significant variation in the resulting collection of synthetic scenes.
3. By precisely customizing and controlling important attributes of the generated scenes, we provide a set of diagnostic benchmarks of previous work on several common computer vision tasks. To our knowledge, ours is the first paper to provide comprehensive diagnostics with respect to algorithm stability and sensitivity to certain scene attributes.
4. We demonstrate the effectiveness of our synthesized scene dataset by advancing the state-of-the-art in the prediction of surface normals and depth from RGB images.

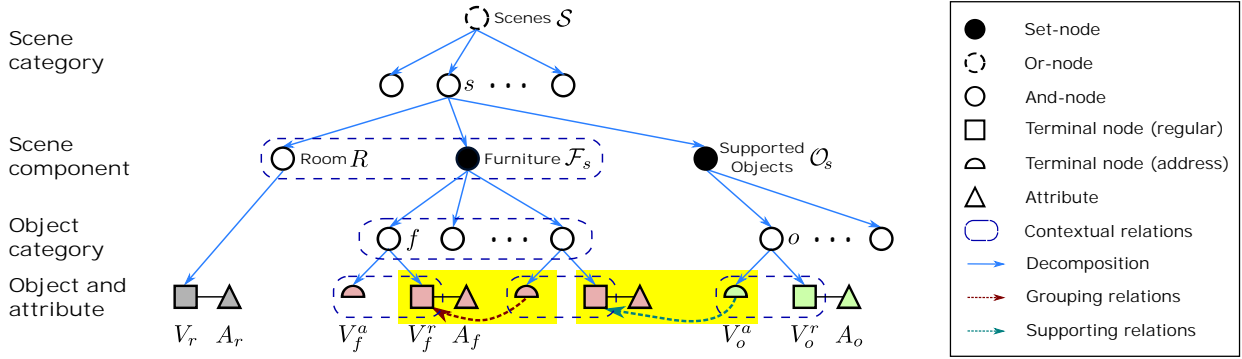

Fig. 2: Scene grammar as an attributed S-AOG. The terminal nodes of the S-AOG are attributed with internal attributes (sizes) and external attributes (positions and orientations). A supported object node is combined by an address terminal node and a regular terminal node, indicating that the object is supported by the furniture pointed to by the address node. If the value of the address node is null, the object is situated on the floor. Contextual relations are defined between walls and furniture, among different furniture pieces, between supported objects and supporting furniture, and for functional groups.

## 2 Representation and Formulation

### 2.1 Representation: Attributed Spatial And-Or Graph

A scene model should be capable of: (i) representing the compositional/hierarchical structure of indoor scenes, and (ii) capturing the rich contextual relationships between different components of the scene. Specifically,

- *Compositional hierarchy* of the indoor scene structure is embedded in a graph representation that models the decomposition into sub-components and the switch among multiple alternative sub-configurations. In general, an indoor scene can first be categorized into different indoor settings (*i.e.*, bedrooms, bathrooms, *etc.*), each of which has a set of walls, furniture, and supported objects. Furniture can be decomposed into functional groups that are composed of multiple pieces of furniture; *e.g.*, a "work" functional group may consist of a desk and a chair.
- *Contextual relations* between pieces of furniture are helpful in distinguishing the functionality of each furniture item and furniture pairs, providing a strong constraint for representing natural indoor scenes. In this paper, we consider four types of contextual relations: (i) relations between furniture pieces and walls; (ii) relations among furniture pieces; (iii) relations between supported objects and their supporting objects (*e.g.*, monitor and desk); and (iv) relations between objects of a functional pair (*e.g.*, sofa and TV).

*Representation:* We represent the hierarchical structure of indoor scenes by an attributed Spatial And-Or Graph (S-AOG), which is a Stochastic Context-Sensitive Grammar (SCSG) with attributes on the terminal nodes. An exam-ple is shown in Figure 2. This representation combines (i) a stochastic context-free grammar (SCFG) and (ii) contextual relations defined on a Markov random field (MRF); *i.e.*, the horizontal links among the terminal nodes. The S-AOG represents the hierarchical decompositions from scenes (top level) to objects (bottom level), whereas contextual relations encode the spatial and functional relations through horizontal links between nodes.

*Definitions:* Formally, an S-AOG is denoted by a 5-tuple: $\mathscr{G} = \langle S, V, R, P, E \rangle$, where $S$ is the root node of the grammar, $V = V_{\text{NT}} \cup V_{\text{T}}$ is the vertex set that includes non-terminal nodes $V_{\text{NT}}$ and terminal nodes $V_{\text{T}}$, $R$ stands for the production rules, $P$ represents the probability model defined on the attributed S-AOG, and $E$ denotes the contextual relations represented as horizontal links between nodes in the same layer.

*Non-terminal Nodes:* The set of non-terminal nodes $V_{\text{NT}} = V^{\text{And}} \cup V^{\text{Or}} \cup V^{\text{Set}}$ is composed of three sets of nodes: *And-nodes* $V^{\text{And}}$ denoting a decomposition of a large entity, *Or-nodes* $V^{\text{Or}}$ representing alternative decompositions, and *Set-nodes* $V^{\text{Set}}$ of which each child branch represents an Or-node on the number of the child object. The Set-nodes are compact representations of nested And-Or relations

*Production Rules:* Corresponding to the three different types of non-terminal nodes, three types of production rules are defined:

1. And rules for an And-node $v \in V^{\text{And}}$ are defined as the deterministic decomposition

$$v \rightarrow u_1 \cdot u_2 \cdot \ldots \cdot u_{n(v)}. \tag{1}$$

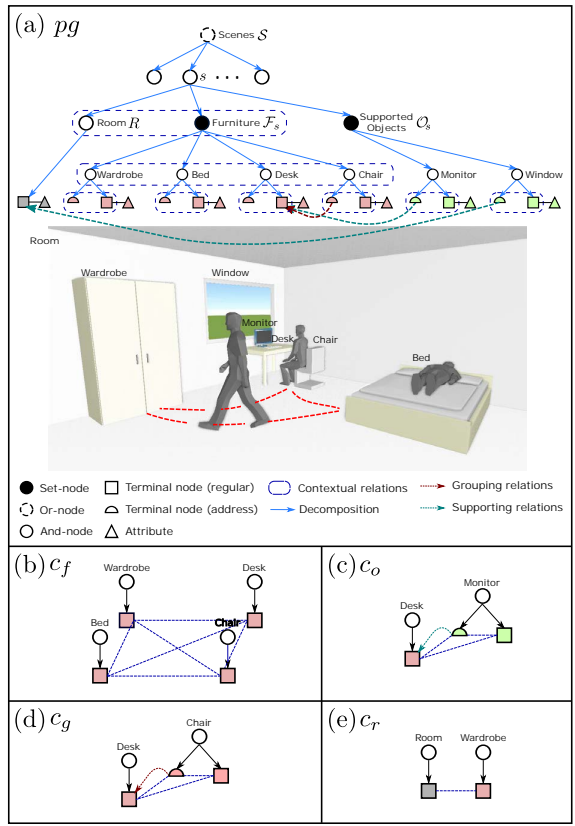

Fig. 3: (a) A simplified example of a parse graph of a bedroom. The terminal nodes of the parse graph form an MRF in the bottom layer. Cliques are formed by the contextual relations projected to the bottom layer. (b)–(e) give an example of the four types of cliques, which represent different contextual relations.

2. Or rules for an Or-node $v \in V^{\mathrm{Or}}$, are defined as the switch

$$v \rightarrow u_1 | u_2 | \dots | u_{n(v)},$$ (2)

with $\rho_1 | \rho_2 | \dots | \rho_{n(v)}$.

3. Set rules for a Set-node $v \in V^{\mathrm{Set}}$ are defined as

$$v \rightarrow (\mathrm{nil}|u_1^1|u_1^2|\dots) \dots (\mathrm{nil}|u_{n(v)}^1|u_{n(v)}^2|\dots),$$ (3)

with $(\rho_{1,0}|\rho_{1,1}|\rho_{1,2}|\dots)\dots(\rho_{n(v),0}|\rho_{n(v),1}|\rho_{n(v),2}|\dots)$, where $u_i^k$ denotes the case that object $u_i$ appears $k$ times, and the probability is $\rho_{i,k}$.

*Terminal Nodes:* The set of terminal nodes can be divided into two types: (i) regular terminal nodes $v \in V_{\mathrm{T}}^{\mathrm{r}}$ representing spatial entities in a scene, with attributes $A$ divided into internal $A_{\mathrm{in}}$ (size) and external $A_{\mathrm{ex}}$ (position and orientation) attributes, and (ii) address terminal nodes $v \in V_{\mathrm{T}}^{\mathrm{a}}$ that point to regular terminal nodes and take values in the set $V_{\mathrm{T}}^{\mathrm{r}} \cup \{\mathrm{nil}\}$. These latter nodes avoid excessively dense graphs by encoding interactions that occur only in a certain context [36].

*Contextual Relations:* The contextual relations $E = E_w \cup E_f \cup E_o \cup E_g$ among nodes are represented by horizontal links in the AOG. The relations are divided into four subsets:

1. relations between furniture pieces and walls $E_w$;
2. relations among furniture pieces $E_f$;
3. relations between supported objects and their supporting objects $E_o$ (*e.g.*, monitor and desk); and
4. relations between objects of a functional pair $E_g$ (*e.g.*, sofa and TV).

Accordingly, the cliques formed in the terminal layer may also be divided into four subsets: $C = C_w \cup C_f \cup C_o \cup C_g$.

Note that the contextual relations of nodes will be inherited from their parents; hence, the relations at a higher level will eventually collapse into cliques $C$ among the terminal nodes. These contextual relations also form an MRF on the terminal nodes. To encode the contextual relations, we define different types of potential functions for different kinds of cliques.

*Parse Tree:* A hierarchical parse tree $pt$ instantiates the S-AOG by selecting a child node for the Or-nodes as well as determining the state of each child node for the Set-nodes. A parse graph $pg$ consists of a parse tree $pt$ and a number of contextual relations $E$ on the parse tree: $pg = (pt, E_{pt})$. Figure 3 illustrates a simple example of a parse graph and four types of cliques formed in the terminal layer.

### 2.2 Probabilistic Formulation

The purpose of representing indoor scenes using an S-AOG is to bring the advantages of compositional hierarchy and contextual relations to bear on the generation of realistic and diverse novel/unseen scene configurations from a learned S-AOG. In this section, we introduce the related probabilistic formulation.

*Prior:* We define the prior probability of a scene configuration generated by an S-AOG using the parameter set $\Theta$. A scene configuration is represented by $pg$, including objects in the scene and their attributes. The prior probability of $pg$ generated by an S-AOG parameterized by $\Theta$ is formulated as a Gibbs distribution,

$$p(pg|\Theta) = \frac{1}{Z} \exp\big(-\mathcal{E}(pg|\Theta)\big)$$ (4)

$$= \frac{1}{Z} \exp\big(-\mathcal{E}(pt|\Theta) - \mathcal{E}(E_{pt}|\Theta)\big),$$ (5)

where $\mathcal{E}(pg|\Theta)$ is the energy function associated with the parse graph, $\mathcal{E}(pt|\Theta)$ is the energy function associated with a parse tree, and $\mathcal{E}(E_{pt}|\Theta)$ is the energy function associated with the contextual relations. Here, $\mathcal{E}(pt|\Theta)$ is defined as combinations of probability distributions with closed-form expressions, and $\mathscr{E}(E_{pt}|\Theta)$ is defined as potential functions relating to the external attributes of the terminal nodes.

*Energy of the Parse Tree:* Energy $\mathscr{E}(pt|\Theta)$ is further decomposed into energy functions associated with different types of non-terminal nodes, and energy functions associated with internal attributes of both regular and address terminal nodes:

$$\mathscr{E}(pt|\Theta) = \underbrace{\sum_{v \in V^{\mathrm{Or}}} \mathscr{E}_{\Theta}^{\mathrm{Or}}(v) + \sum_{v \in V^{\mathrm{Set}}} \mathscr{E}_{\Theta}^{\mathrm{Set}}(v)}_{\text{non-terminal nodes}} + \underbrace{\sum_{v \in V_T^A} \mathscr{E}_{\Theta}^{A_{\mathrm{in}}}(v)}_{\text{terminal nodes}}, \quad (6)$$

where the choice of child node of an Or-node $v \in V^{\mathrm{Or}}$ follows a multinomial distribution, and each child branch of a Set-node $v \in V^{\mathrm{Set}}$ follows a Bernoulli distribution. Note that the And-nodes are deterministically expanded; hence, (6) lacks an energy term for the And-nodes. The internal attributes $A_{\mathrm{in}}$ (size) of terminal nodes follows a nonparametric probability distribution learned via kernel density estimation.

*Energy of the Contextual Relations:* The energy $\mathscr{E}(E_{pt}|\Theta)$ is described by the probability distribution

$$p(E_{pt}|\Theta) = \frac{1}{Z} \exp\left(-\mathscr{E}(E_{pt}|\Theta)\right) \quad (7)$$

$$= \prod_{c \in C_w} \phi_w(c) \prod_{c \in C_f} \phi_f(c) \prod_{c \in C_o} \phi_o(c) \prod_{c \in C_g} \phi_g(c), \quad (8)$$

which combines the potentials of the four types of cliques formed in the terminal layer. The potentials of these cliques are computed based on the external attributes of regular terminal nodes:

1. Potential function $\phi_w(c)$ is defined on relations between walls and furniture (Figure 3b). A clique $c \in C_w$ includes a terminal node representing a piece of furniture $f$ and the terminal nodes representing walls $\{w_i\}$: $c = \{f, \{w_i\}\}$. Assuming pairwise object relations, we have

$$\phi_w(c) = \frac{1}{Z} \exp\left(-\lambda_w \cdot \left\langle \underbrace{\sum_{w_i \ne w_j} l_{\mathrm{con}}(w_i, w_j)}_{\text{constraint between walls}}, \right.\right.$$
$$\left.\left. \underbrace{\sum_{w_i} [l_{\mathrm{dis}}(f, w_i) + l_{\mathrm{ori}}(f, w_i)]}_{\text{constraint between walls and furniture}} \right\rangle\right), \quad (9)$$

where $\lambda_w$ is a weight vector, and $l_{\mathrm{con}}$, $l_{\mathrm{dis}}$, $l_{\mathrm{ori}}$ are three different cost functions:

   (a) The cost function $l_{\mathrm{con}}(w_i, w_j)$ defines the consistency between the walls; *i.e.*, adjacent walls should be connected, whereas opposite walls should have the same size. Although this term is repeatedly computed in different cliques, it is usually zero as the walls are enforced to be consistent in practice.

   (b) The cost function $l_{\mathrm{dis}}(x_i, x_j)$ defines the geometric distance compatibility between two objects
$$l_{\mathrm{dis}}(x_i, x_j) = |d(x_i, x_j) - \bar{d}(x_i, x_j)|, \quad (10)$$
   where $d(x_i, x_j)$ is the distance between object $x_i$ and $x_j$, and $\bar{d}(x_i, x_j)$ is the mean distance learned from all the examples.

   (c) Similarly, the cost function $l_{\mathrm{ori}}(x_i, x_j)$ is defined as
$$l_{\mathrm{ori}}(x_i, x_j) = |\theta(x_i, x_j) - \bar{\theta}(x_i, x_j)|, \quad (11)$$
   where $\theta(x_i, x_j)$ is the distance between object $x_i$ and $x_j$, and $\bar{\theta}(x_i, x_j)$ is the mean distance learned from all the examples. This term represents the compatibility between two objects in terms of their relative orientations.

2. Potential function $\phi_f(c)$ is defined on relations between pieces of furniture (Figure 3c). A clique $c \in C_f$ includes all the terminal nodes representing a piece of furniture: $c = \{f_i\}$. Hence,

$$\phi_f(c) = \frac{1}{Z} \exp\left(-\lambda_c \sum_{f_i \ne f_j} l_{\mathrm{occ}}(f_i, f_j)\right), \quad (12)$$

where the cost function $l_{\mathrm{occ}}(f_i, f_j)$ defines the compatibility of two pieces of furniture in terms of occluding accessible space

$$l_{\mathrm{occ}}(f_i, f_j) = \max(0, 1 - d(f_i, f_j)/d_{\mathrm{acc}}). \quad (13)$$

3. Potential function $\phi_o(c)$ is defined on relations between a supported object and the furniture piece that supports it (Figure 3d). A clique $c \in C_o$ consists of a supported object terminal node $o$, the address node $a$ connected to the object, and the furniture terminal node $f$ pointed to by the address node $c = \{f, a, o\}$:

$$\phi_o(c) = \frac{1}{Z} \exp\left(-\lambda_o \cdot \left\langle l_{\mathrm{pos}}(f, o), l_{\mathrm{ori}}(f, o), l_{\mathrm{add}}(a) \right\rangle\right), (14)$$

which incorporates three different cost functions. The cost function $l_{\mathrm{ori}}(f, o)$ has been defined for potential function $\phi_w(c)$, and the two new cost functions are as follows:

   (a) The cost function $l_{\mathrm{pos}}(f, o)$ defines the relative position of the supported object $o$ to the four boundaries of the bounding box of the supporting furniture $f$:
$$l_{\mathrm{pos}}(f, o) = \sum_i l_{\mathrm{dis}}(f_{\mathrm{face}_i}, o). \quad (15)$$

   (b) The cost term $l_{\mathrm{add}}(a)$ is the negative log probability of an address node $v \in V_T^a$, which is regarded as a certain regular terminal node and follows a multinomial distribution.

4. Potential function $\phi_g(c)$ is defined for furniture in the same functional group (Figure 3d). A clique $c \in C_g$ consists of terminal nodes representing furniture in a functional group g: $c = \{f_i^g\}$:

$$\phi_g(c) = \frac{1}{Z} \exp\left(-\sum_{f_i^g \ne f_j^g} \lambda_g \cdot \left\langle l_{\mathrm{dis}}(f_i^g, f_j^g), l_{\mathrm{ori}}(f_i^g, f_j^g) \right\rangle\right). (16)$$

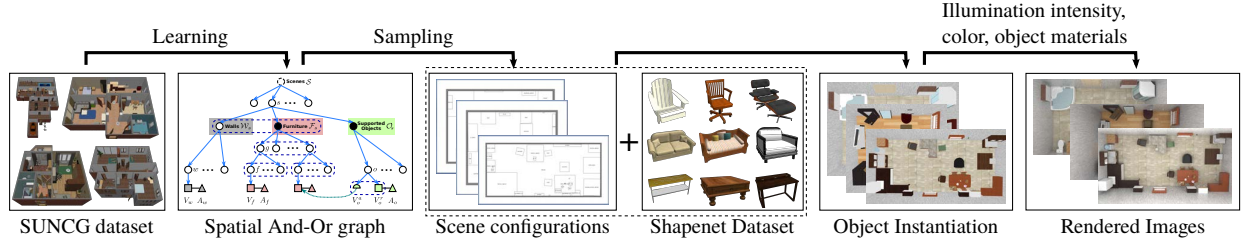

Fig. 4: The learning-based pipeline for synthesizing images of indoor scenes.

## 3 Learning, Sampling, and Synthesis

Before we introduce in Section 3.1 the algorithm for learning the parameters associated with an S-AOG, note that our configurable scene synthesis pipeline includes the following components:

- A sampling algorithm based on the learned S-AOG for synthesizing realistic scene geometric configurations. This sampling algorithm controls the size of the individual objects as well as their pair-wise relations. More complex relations are recursively formed using pair-wised relations. The details are found in Section 3.2.
- An attribute assignment process, which sets different material attributes to each object part, as well as various camera parameters and illuminations of the environment. The details are found in Section 3.4.

The above two components are the essence of *configurable* scene synthesis; the first generates the structure of the scene while the second controls its detailed attributes. In between these two components, a scene instantiation process is applied to generate a 3D mesh of the scene based on the sampled scene layout. This step is described in Section 3.3. Figure 4 illustrates the pipeline.

### 3.1 Learning the S-AOG

The parameters $\Theta$ of a probability model can be learned in a supervised way from a set of $N$ observed parse trees $\{pt_n\}_{n=1,\dots,N}$ by maximum likelihood estimation (MLE):

$$\Theta^* = \arg\max_{\Theta} \prod_{n=1}^{N} p(pt_n|\Theta). \qquad (17)$$

We now describe how to learn all the parameters $\Theta$, with the focus on learning the weights of the loss functions.

*Weights of the Loss Functions:* Recall that the probability distribution of cliques formed in the terminal layer is given by (8); *i.e.*,

$$p(E_{pt}|\Theta) = \frac{1}{Z}\exp\left(-\mathscr{E}(E_{pt}|\Theta)\right) \qquad (18)$$

$$= \frac{1}{Z}\exp\left(-\lambda \cdot l(E_{pt})\right), \qquad (19)$$

where $\lambda$ is the weight vector and $l(E_{pt})$ is the loss vector given by the four different types of potential functions. To learn the weight vector, the traditional MLE maximizes the average log-likelihood

$$\mathscr{L}(E_{pt}|\Theta) = \frac{1}{N}\sum_{n=1}^{N}\log p(E_{pt_n}|\Theta) \qquad (20)$$

$$= -\frac{1}{N}\sum_{n=1}^{N}\lambda \cdot l(E_{pt_n}) - \log Z, \qquad (21)$$

usually by energy gradient descent:

$$\frac{\partial \mathscr{L}(E_{pt}|\Theta)}{\partial \lambda}$$

$$= -\frac{1}{N}\sum_{n=1}^{N} l(E_{pt_n}) - \frac{\partial \log Z}{\partial \lambda} \qquad (22)$$

$$= -\frac{1}{N}\sum_{n=1}^{N} l(E_{pt_n}) - \frac{\partial \log \sum_{pt}\exp\left(-\lambda \cdot l(E_{pt})\right)}{\partial \lambda} \qquad (23)$$

$$= -\frac{1}{N}\sum_{n=1}^{N} l(E_{pt_n}) + \sum_{pt}\frac{1}{Z}\exp\left(-\lambda \cdot l(E_{pt})\right)l(E_{pt}) \qquad (24)$$

$$= -\frac{1}{N}\sum_{n=1}^{N} l(E_{pt_n}) + \frac{1}{\tilde{N}}\sum_{\tilde{n}=1}^{\tilde{N}} l(E_{pt_{\tilde{n}}}), \qquad (25)$$

where $\{E_{pt_{\tilde{n}}}\}_{\tilde{n}=1,\dots,\tilde{N}}$ is the set of synthesized examples from the current model.

Unfortunately, it is computationally infeasible to sample a Markov chain that turns into an *equilibrium distribution* at every iteration of gradient descent. Hence, instead of waiting for the Markov chain to converge, we adopt the contrastive divergence (CD) learning that follows the gradient of the difference of two divergences [55]:

$$\mathrm{CD}_{\tilde{N}} = \mathrm{KL}(p_0||p_\infty) - \mathrm{KL}(p_{\tilde{n}}||p_\infty), \qquad (26)$$

where $\mathrm{KL}(p_0||p_\infty)$ is the Kullback-Leibler divergence between the data distribution $p_0$ and the model distribution $p_\infty$, and $p_{\tilde{n}}$ is the distribution obtained by a Markov chain started at the data distribution and run for a small number $\tilde{n}$ of steps (*e.g.*, $\tilde{n} = 1$). Contrastive divergence learning has been applied effectively in addressing various problems, most notably in the context of Restricted Boltzmann Machines [56].

Both theoretical and empirical evidence corroborates its efficiency and very small bias [13]. The gradient of the contrastive divergence is given by:

$$\frac{\partial \mathrm{CD}_{\tilde{N}}}{\partial \lambda} = \frac{1}{N}\sum_{n=1}^{N} l(E_{pt_n}) - \frac{1}{\tilde{N}}\sum_{n=1}^{\tilde{N}} l(E_{pt_{\tilde{n}}}) \qquad (27)$$

$$- \frac{\partial p_{\tilde{n}}}{\partial \lambda}\frac{\partial \mathrm{KL}(p_{\tilde{n}}||p_{\infty})}{\partial p_{\tilde{n}}}.$$

Extensive simulations [55] showed that the third term can be safely ignored since it is small and seldom opposes the resultant of the other two terms.

Finally, the weight vector is learned by gradient descent computed by generating a small number $\tilde{n}$ of examples from the Markov chain

$$\lambda_{t+1} = \lambda_t - \eta_t \frac{\partial \mathrm{CD}_{\tilde{N}}}{\partial \lambda} \qquad (28)$$

$$= \lambda_t + \eta_t \left( \frac{1}{\tilde{N}}\sum_{\tilde{n}=1}^{\tilde{N}} l(E_{pt_{\tilde{n}}}) - \frac{1}{N}\sum_{n=1}^{N} l(E_{pt_n}) \right). \qquad (29)$$

*Or-nodes and Address-nodes:* The MLE of the branching probabilities of Or-nodes and address terminal nodes is simply the frequency of each alternative choice [152]:

$$\rho_i = \frac{\#(v \to u_i)}{\sum\limits_{j=1}^{n(v)} \#(v \to u_j)}; \qquad (30)$$

however, the samples we draw from the distributions will rarely cover all possible terminal nodes to which an address node is pointing, since there are many unseen but plausible configurations. For instance, an apple can be put on a chair, which is semantically and physically plausible, but the training examples are highly unlikely to include such a case. Inspired by the Dirichlet process, we address this issue by altering the MLE to include a small probability $\alpha$ for all branches:

$$\rho_i = \frac{\#(v \to u_i) + \alpha}{\sum\limits_{j=1}^{n(v)} (\#(v \to u_j) + \alpha)}. \qquad (31)$$

*Set-nodes:* Similarly, for each child branch of the Set-nodes, we use the frequency of samples as the probability, if it is non-zero, otherwise we set the probability to $\alpha$. Based on the common practice—*e.g.*, choosing the probability of joining a new table in the Chinese restaurant process [1]—we set $\alpha$ to have probability 1.

*Parameters:* To learn the S-AOG for sampling purposes, we collect statistics using the SUNCG dataset [114], which contains over 45K different scenes with manually created realistic room and furniture layouts. We collect the statistics of room types, room sizes, furniture occurrences, furniture sizes, relative distances and orientations between furniture and walls, furniture affordance, grouping occurrences, and supporting relations.

The parameters of the loss functions are learned from the constructed scenes by computing the statistics of relative distances and relative orientations between different objects.

The grouping relations are manually defined (*e.g.*, nightstands are associated with beds, chairs are associated with desks and tables). We examine each pair of furniture pieces in the scene, and a pair is regarded as a group if the distance of the pieces is smaller than a threshold (*e.g.*, 1m). The probability of occurrence is learned as a multinomial distribution. The supporting relations are automatically discovered from the dataset by computing the vertical distance between pairs of objects and checking if one bounding polygon contains another.

The distribution of object size among all the furniture and supported objects is learned from the 3D models provided by the ShapeNet dataset [15] and the SUNCG dataset [114]. We first extracted the size information from the 3D models, and then fitted a non-parametric distribution using kernel density estimation. Not only is this more accurate than simply fitting a trivariate normal distribution, but it is also easier to sample from.

### 3.2 Sampling Scene Geometry Configurations

Based on the learned S-AOG, we sample scene configurations (parse graphs) based on the prior probability $p(pg|\Theta)$ using a Markov Chain Monte Carlo (MCMC) sampler. The sampling process comprises two major steps:

1. Top-down sampling of the parse tree structure *pt* and internal attributes of objects. This step selects a branch for each Or-node and chooses a child branch for each Set-node. In addition, internal attributes (sizes) of each regular terminal node are also sampled. Note that this can be easily done by sampling from closed-form distributions.

2. MCMC sampling of the external attributes (positions and orientations) of objects as well as the values of the address nodes. Samples are proposed by Markov chain dynamics, and are taken after the Markov chain converges to the prior probability. These attributes are constrained by multiple potential functions, hence it is difficult to sample directly from the true underlying probability distribution.

Algorithm 1 overviews the sampling process. Some qualitative results are shown in Figure 5.

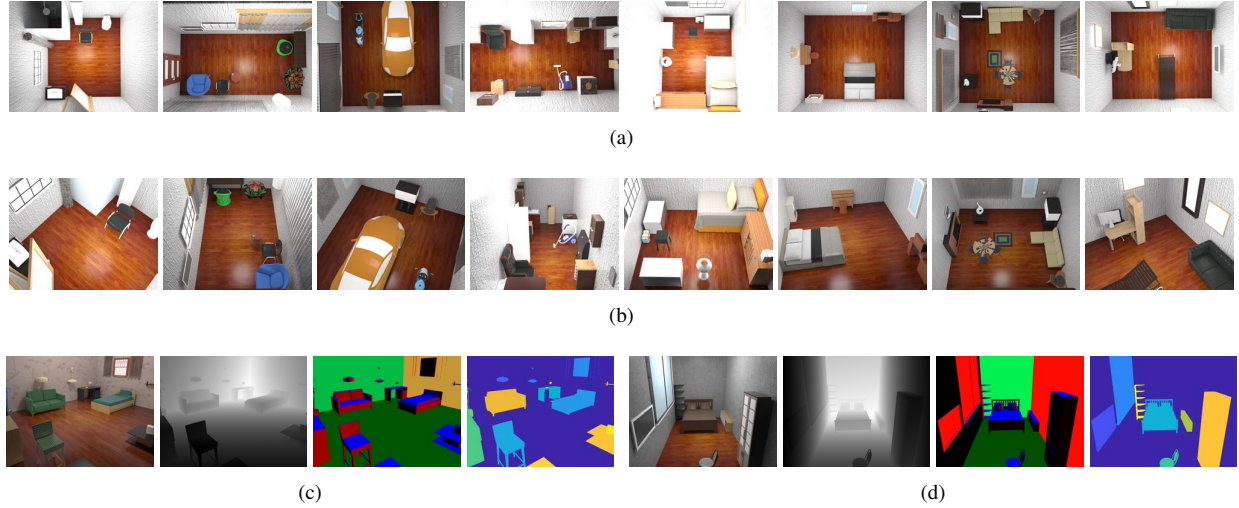

(a)

(b)

(c)                                                        (d)

Fig. 5: Qualitative results in different types of scenes using default attributes of object materials, illumination conditions, and camera parameters; (a) overhead view; (b) random view. (c) (d) Additional examples of two bedrooms, with (from left) image, with corresponding depth map, surface normal map, and semantic segmentation.

---

**Algorithm 1:** Sampling Scene Configurations

**Input** : Attributed S-AOG $\mathscr{G}$
        Landscape parameter $\beta$
        sample number $n$
**Output**: Synthesized room layouts $\{pg_i\}_{i=1,\ldots,n}$

1 **for** $i = 1$ **to** $n$ **do**
2     Sample the child nodes of the Set nodes and Or nodes
      from $\mathscr{G}$ directly to obtain the structure of $pg_i$.
3     Sample the sizes of room, furniture $f$, and objects $o$ in $pg_i$
      directly.
4     Sample the address nodes $V^a$.
5     Randomly initialize positions and orientations of furniture
      $f$ and objects $o$ in $pg_i$.
6     $iter = 0$
7     **while** $iter < iter_{\max}$ **do**
8         Propose a new move and obtain proposal $pg_i'$.
9         Sample $u \sim \text{unif}(0,1)$.
10       **if** $u < \min(1, \exp(\beta(\mathscr{E}(pg_i|\Theta) - \mathscr{E}(pg_i'|\Theta))))$ **then**
11         $pg_i = pg_i'$
12       **end**
13       $iter += 1$
14     **end**
15 **end**

---

*Markov Chain Dynamics:* To propose moves, four types of Markov chain dynamics, $q_i, i = 1, 2, 3, 4$, are designed to be chosen randomly with certain probabilities. Specifically, the dynamics $q_1$ and $q_2$ are diffusion, while $q_3$ and $q_4$ are reversible jumps:

1. *Translation of Objects.* Dynamic $q_1$ chooses a regular terminal node and samples a new position based on the current position of the object,

$$\text{pos} \rightarrow \text{pos} + \delta\text{pos}, \tag{32}$$

where $\delta$pos follows a bivariate normal distribution.

2. *Rotation of Objects.* Dynamic $q_2$ chooses a regular terminal node and samples a new orientation based on the current orientation of the object,

$$\theta \rightarrow \theta + \delta\theta, \tag{33}$$

where $\delta\theta$ follows a normal distribution.

3. *Swapping of Objects.* Dynamic $q_3$ chooses two regular terminal nodes and swaps the positions and orientations of the objects.

4. *Swapping of Supporting Objects.* Dynamic $q_4$ chooses an address terminal node and samples a new regular furniture terminal node pointed to. We sample a new 3D location $(x, y, z)$ for the supported object:

- Randomly sample $x = u_x w_p$, where $u_x \sim \text{unif}(0,1)$, and $w_p$ is the width of the supporting object.
- Randomly sample $y = u_y l_p$, where $u_y \sim \text{unif}(0,1)$, and $l_p$ is the length of the supporting object.
- The height $z$ is simply the height of the supporting object.

Adopting the Metropolis-Hastings algorithm, a newly proposed parse graph $pg'$ is accepted according to the following acceptance probability:

$$\alpha(pg'|pg, \Theta) = \min(1, \frac{p(pg'|\Theta)p(pg|pg')}{p(pg|\Theta)p(pg'|pg)}) \tag{34}$$

$$= \min(1, \frac{p(pg'|\Theta)}{p(pg|\Theta)}) \tag{35}$$

$$= \min(1, \exp(\mathscr{E}(pg|\Theta) - \mathscr{E}(pg'|\Theta))). \tag{36}$$

The proposal probabilities cancel since the proposed moves are symmetric in probability.

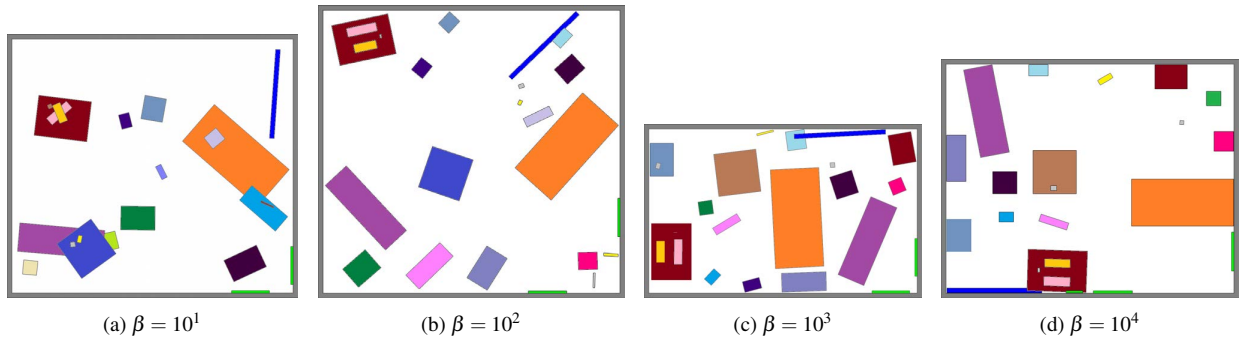

(a) $\beta = 10^1$      (b) $\beta = 10^2$      (c) $\beta = 10^3$      (d) $\beta = 10^4$

Fig. 6: Synthesis for different values of $\beta$. Each image shows a typical configuration sampled from a Markov chain.

*Convergence:* To test if the Markov chain has converged to the prior probability, we maintain a histogram of the energy of the last $w$ samples. When the difference between two histograms separated by $s$ sampling steps is smaller than a threshold $\varepsilon$, the Markov chain is considered to have converged.

*Tidiness of Scenes:* During the sampling process, a typical state is drawn from the distribution. We can easily control the level of tidiness of the sampled scenes by adding an extra parameter $\beta$ to control the landscape of the prior distribution:

$$p(pg|\Theta) = \frac{1}{Z}\exp\left(-\beta\mathscr{E}(pg|\Theta)\right). \quad (37)$$

Some examples are shown in Figure 6.

Note that the parameter $\beta$ is analogous to but differs from the temperature in simulated annealing optimization—the temperature in simulated annealing is time-variant; *i.e.*, it changes during the simulated annealing process. In our model, we simulate a Markov chain under one specific $\beta$ to get typical samples at a certain level of tidiness. When $\beta$ is small, the distribution is "smooth"; *i.e.*, the differences between local minima and local maxima are small.

### 3.3 Scene Instantiation using 3D Object Datasets

Given a generated 3D scene layout, the 3D scene is instantiated by assembling objects into it using 3D object datasets. We incorporate both the ShapeNet dataset [14] and the SUNCG dataset [114] as our 3D model dataset. Scene instantiation includes the following five steps:

1. For each object in the scene layout, find the model that has the closest length/width ratio to the dimension specified in the scene layout.
2. Align the orientations of the selected models according to the orientation specified in the scene layout.

3. Transform the models to the specified positions, and scale the models according to the generated scene layout.
4. Since we fit only the length and width in Step 1, an extra step to adjust the object position along the gravity direction is needed to eliminate floating models and models that penetrate into one another.
5. Add the floor, walls, and ceiling to complete the instantiated scene.

### 3.4 Scene Attribute Configurations

As we generate scenes in a forward manner, our pipeline enables the precise customization and control of important attributes of the generated scenes. Some configurations are shown in Figure 7. The rendered images are determined by combinations of the following four factors:

- Illuminations, including the number of light sources, and the light source positions, intensities, and colors.
- Material and textures of the environment; *i.e.*, the walls, floor, and ceiling.
- Cameras, such as fisheye, panorama, and Kinect cameras, have different focal lengths and apertures, yielding dramatically different rendered images. By virtue of physics-based rendering, our pipeline can even control the F-stop and focal distance, resulting in different depths of field.
- Different object materials and textures will have various properties, represented by roughness, metallicness, and reflectivity.

## 4 Photorealistic Scene Rendering

We adopt Physics-Based Rendering (PBR) [92] to generate the photorealistic 2D images. PBR has become the industry standard in computer graphics applications in recent years, and it has been widely adopted for both offline and

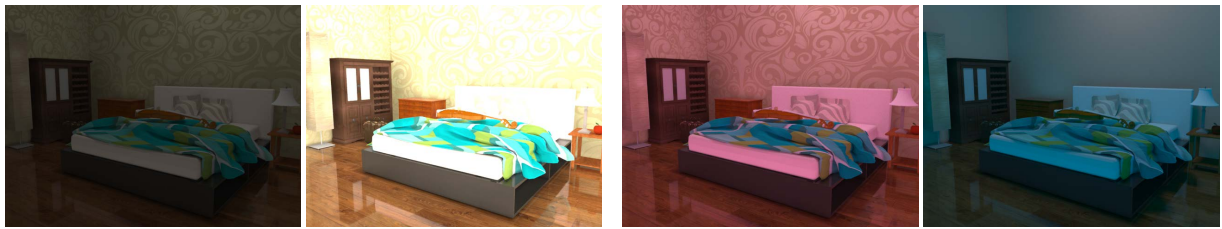

(a) Illumination intensity: half and double

(b) Illumination color: purple and blue

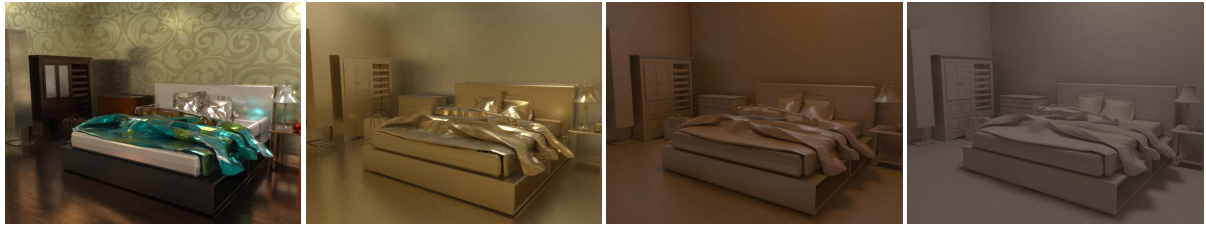

(c) Different object materials: metal, gold, chocolate, and clay

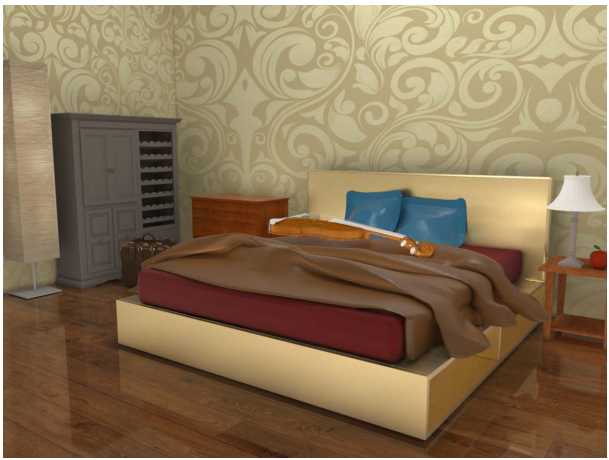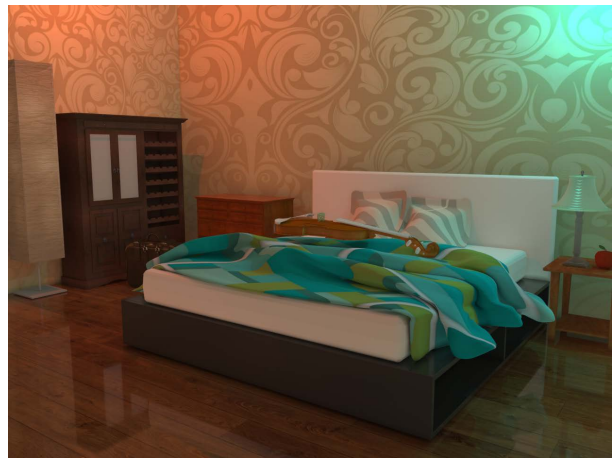

(d) Different materials in each object part

(e) Multiple light sources

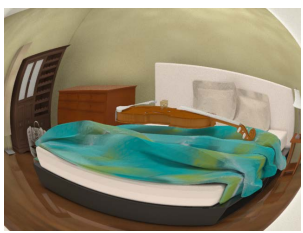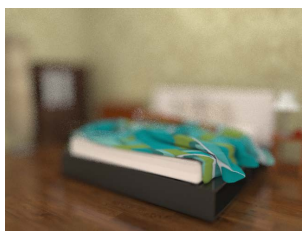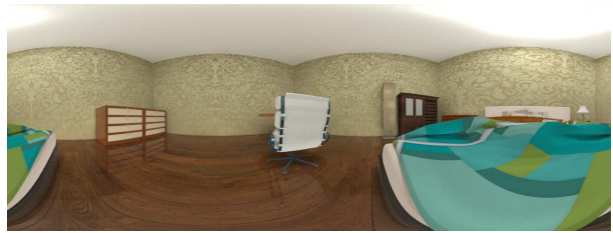

(f) Fish eye lens        (g) Image with depth of field        (h) Panorama image

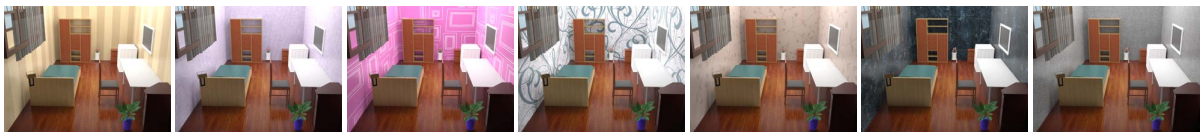

(i) Different background materials affect the rendering results

Fig. 7: We can configure the scene with different (a) illumination intensities, (b) illumination colors, and (c) materials, (d) even on each object part. We can also control (e) the number of light source and their positions, (f) camera lenses (*e.g.*, fish eye), (g) depths of field, or (h) render the scene as a panorama for virtual reality and other virtual environments. (i) Seven different background wall textures. Note how the background affects the overall illumination.

real-time rendering. Unlike traditional rendering techniques where heuristic shaders are used to control how light is scattered by a surface, PBR simulates the physics of real-world light by computing the bidirectional scattering distribution function (BSDF) [6] of the surface.

*Formulation:* Following the law of conservation of energy, PBR solves the rendering equation for the total spectral radiance of outgoing light $L_o(\mathbf{x}, \mathbf{w})$ in direction $\mathbf{w}$ from point $\mathbf{x}$ on a surface as

$$L_o(\mathbf{x}, \mathbf{w}) = L_e(\mathbf{x}, \mathbf{w}) \qquad\qquad (38)$$
$$+ \int_{\Omega} f_r(\mathbf{x}, \mathbf{w}', \mathbf{w}) L_i(\mathbf{x}, \mathbf{w}')(-\mathbf{w}' \cdot \mathbf{n}) \, d\mathbf{w}',$$

where $L_e$ is the emitted light (from a light source), $\Omega$ is the unit hemisphere uniquely determined by $\mathbf{x}$ and its normal, $f_r$ is the bidirectional reflectance distribution function (BRDF), $L_i$ is the incoming light from direction $\mathbf{w}'$, and $\mathbf{w}' \cdot \mathbf{n}$ accounts for the attenuation of the incoming light.

*Advantages:* In path tracing, the rendering equation is often computed using Monte Carlo methods. Contrasting what happens in the real world, the paths of photons in a scene are traced backwards from the camera (screen pixels) to the light sources. Objects in the scene receive illumination contributions as they interact with the photon paths. By computing both the reflected and transmitted components of rays in a physically accurate way, while conserving energies and obeying refraction equations, PBR photorealistically renders shadows, reflections, and refractions, thereby synthesizing superior levels of visual detail compared to other shading techniques. Note PBR describes a shading process and does not dictate how images are rasterized in screen space. We use the *Mantra*® PBR engine to render synthetic image data with ray tracing for its accurate calculation of illumination and shading as well as its physically intuitive parameter configurability.

Indoor scenes are typically closed rooms. Various reflective and diffusive surfaces may exist throughout the space. Therefore, the effect of secondary rays is particularly important in achieving realistic illumination. PBR robustly samples both direct illumination contributions on surfaces from light sources and indirect illumination from rays reflected and diffused by other surfaces. The BSDF shader on a surface manages and modifies its color contribution when hit by a secondary ray. Doing so results in more secondary rays being sent out from the surface being evaluated. The reflection limit (the number of times a ray can be reflected) and the diffuse limit (the number of times diffuse rays bounce on surfaces) need to be chosen wisely to balance the final image quality and rendering time. Decreasing the number of indirect illumination samples will likely yield a nice rendering time reduction, but at the cost of significantly diminished visual realism.

*Rendering Time vs Rendering Quality:* In summary, we use the following control parameters to adjust the quality and speed of rendering:

- *Baseline pixel samples.* This is the minimum number of rays sent per pixel. Each pixel is typically divided evenly in both directions. Common values for this parameter are $3 \times 3$ and $5 \times 5$. The higher pixel sample counts are usually required to produce motion blur and depth of field effects.
- *Noise level.* Different rays sent from each pixel will not yield identical paths. This parameter determines the maximum allowed variance among the different results. If necessary, additional rays (in addition to baseline pixel sample count) will be generated to decrease the noise.
- *Maximum additional rays.* This parameter is the upper limit of the additional rays sent for satisfying the noise level.
- *Bounce limit.* The maximum number of secondary ray bounces. We use this parameter to restrict both diffuse and reflected rays. Note that in PBR the diffuse ray is one of the most significant contributors to realistic global illumination, while the other parameters are more important in controlling the Monte Carlo sampling noise.

Table 1 summarizes our analysis of how these parameters affect the rendering time and image quality.

## 5 Experiments

In this section, we demonstrate the usefulness of the generated synthetic indoor scenes from two perspectives:

1. Improving state-of-the-art computer vision models by training with our synthetic data. We showcase our results on the task of normal prediction and depth prediction from a single RGB image, demonstrating the potential of using the proposed dataset.
2. Benchmarking common scene understanding tasks with configurable object attributes and various environments, which evaluates the stabilities and sensitivities of the algorithms, providing directions and guidelines for their further improvement in various vision tasks.

The reported results use the reference parameters indicated in Table 1. Using the Mantra renderer, we found that choosing parameters to produce lower-quality rendering does not provide training images that suffice to outperform the state-of-the-art methods using the experimental setup described below.

### 5.1 Normal Estimation

Estimating surface normals from a single RGB image is an essential task in scene understanding, since it provides important information in recovering the 3D structures of scenes. We train a neural network using our synthetic data to demonstrate that the perfect per-pixel ground truth generated using our pipeline may be utilized to improve upon the state-of-the-art performance on this specific scene understanding task. Using the fully convolutional network model described by Zhang *et al.* [146], we compare the normal estimation results given by models trained under two different protocols: (i) the network is directly trained and tested on the NYU-Depth V2 dataset and (ii) the network is first pretrained using our synthetic data, then fine-tuned and tested on NYU-Depth V2.

Table 1: Comparisons of rendering time vs quality. The first column tabulates the reference number and rendering results used in this paper, the second column lists all the criteria, and the remaining columns present comparative results. The color differences between the reference image and images rendered with various parameters are measured by the LAB Delta E standard [110] tracing back to Helmholtz and Hering [2, 125].

| Reference | Criteria | Comparisons | | | | | | | |
|---|---|---|---|---|---|---|---|---|---|
| $3 \times 3$ | Baseline pixel samples | $2 \times 2$ | $1 \times 1$ | $3 \times 3$ | $3 \times 3$ | $3 \times 3$ | $3 \times 3$ | $3 \times 3$ | $3 \times 3$ |
| 0.001 | Noise level | 0.001 | 0.001 | **0.01** | **0.1** | 0.001 | 0.001 | 0.001 | 0.001 |
| 22 | Maximum additional rays | 22 | 22 | 22 | 22 | **10** | **3** | 22 | 22 |
| 6 | Bounce limit | 6 | 6 | 6 | 6 | 6 | 6 | **3** | **1** |
| 203 | Time (seconds) | 131 | 45 | 196 | 30 | 97 | 36 | 198 | 178 |
| 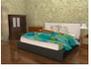 | LAB Delta E difference | 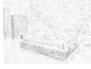 | 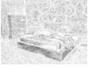 | 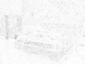 | 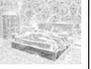 | 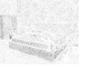 | 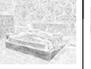 | 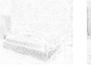 | 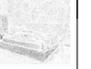 |

Table 2: Performance of normal estimation for the NYU-Depth V2 dataset with different training protocols.

| Pre-train | Fine-tune | Mean↓ | Median↓ | 11.25°↑ | 22.5°↑ | 30.0°↑ |
|---|---|---|---|---|---|---|
| | NYUv2 | 27.30 | 21.12 | 27.21 | 52.61 | 64.72 |
| | Eigen | 22.2 | 15.3 | 38.6 | 64.0 | 73.9 |
| [146] | NYUv2 | 21.74 | 14.75 | 39.37 | 66.25 | 76.06 |
| ours+ [146] | NYUv2 | **21.47** | **14.45** | **39.84** | **67.05** | **76.72** |

Following the standard protocol [3, 35], we evaluate a per-pixel error over the entire dataset. To evaluate the prediction error, we computed the mean, median, and RMSE of angular error between the predicted normals and ground truth normals. Prediction accuracy is given by calculating the fraction of pixels that are correct within a threshold $t$, where $t = 11.25°$, $22.5°$, and $30.0°$. Our experimental results are summarized in Table 2. By utilizing our synthetic data, the model achieves better performance. From the visualized results in Figure 8, we can see that the error mainly accrues in the area where the ground truth normal map is noisy. We argue that the reason is partly due to sensor noise or sensing distance limit. Our results indicate the importance of having perfect per-pixel ground truth for training and evaluation.

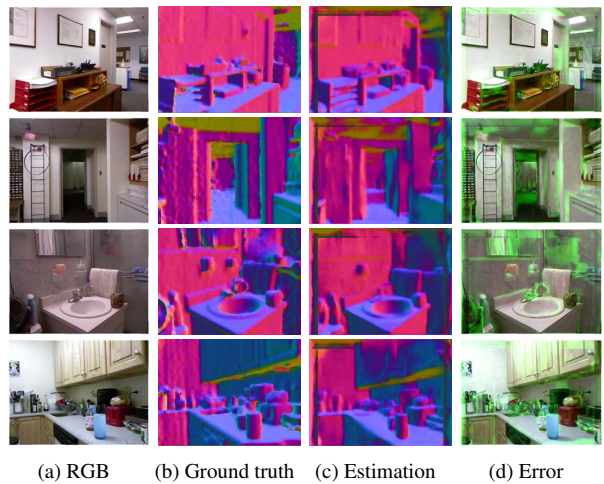

(a) RGB     (b) Ground truth     (c) Estimation     (d) Error

Fig. 8: Examples of normal estimation results predicted by the model trained with our synthetic data.

### 5.2 Depth Estimation

Depth estimation is a fundamental and challenging problem in computer vision that is broadly applicable in scene understanding, 3D modeling, and robotics. In this task, the algorithms output a depth image based on a single RGB input image.

To demonstrate the efficacy of our synthetic data, we compare the depth estimation results provided by models trained following protocols similar to those we used in normal estimation with the network in [71]. To perform a quantitative evaluation, we used the metrics applied in previous work [29]:

- Abs relative error: $\frac{1}{N} \sum_p |d_p - d_p^{gt}| / d_p^{gt}$,
- Square relative difference: $\frac{1}{N} \sum_p |d_p - d_p^{gt}|^2 / d_p^{gt}$,
- Average $\log_{10}$ error: $\frac{1}{N} \sum_x |\log_{10}(d_p) - \log_{10}(d_p^{gt})|$,
- RMSE : $\left( \frac{1}{N} \sum_x |d_p - d_p^{gt}|^2 \right)^{1/2}$,

Table 3: Depth estimation performance on the NYU-Depth V2 dataset with different training protocols.

| Pre-Train | Fine-Tune | Error | | | | | Accuracy | | |
|---|---|---|---|---|---|---|---|---|---|
| | | Abs Rel | Sqr Rel | Log10 | RMSE (linear) | RMSE (log) | $\delta < 1.25$ | $\delta < 1.25^2$ | $\delta < 1.25^3$ |
| NYUv2 | - | 0.233 | 0.158 | 0.098 | 0.831 | 0.117 | 0.605 | 0.879 | 0.965 |
| Ours | - | 0.241 | 0.173 | 0.108 | 0.842 | 0.125 | 0.612 | 0.882 | 0.966 |
| Ours | NYUv2 | **0.226** | **0.152** | **0.090** | **0.820** | **0.108** | **0.616** | **0.887** | **0.972** |

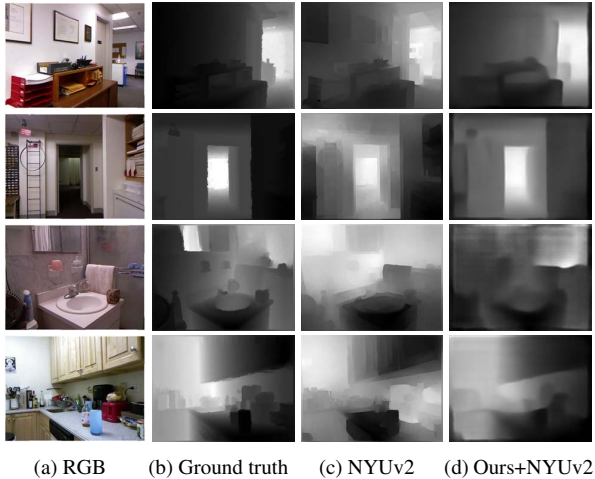

(a) RGB        (b) Ground truth        (c) NYUv2        (d) Ours+NYUv2

Fig. 9: Examples of depth estimation results predicted by the model trained with our synthetic data.

- Log RMSE: $\left( \frac{1}{N} \sum_x \left| \log(d_p) - \log(d_p^{gt}) \right|^2 \right)^{1/2}$,
- Threshold: % of $d_p$ s.t. $\max \left( d_p/d_p^{gt}, d_p^{gt}/d_p \right) <$ threshold, where $d_p$ and $d_p^{gt}$ are the predicted depths and the ground truth depths, respectively, at the pixel indexed by $p$ and $N$ is the number of pixels in all the evaluated images. The first five metrics capture the error calculated over all the pixels; lower values are better. The threshold criteria capture the estimation accuracy; higher values are better.

Table 3 summarizes the results. We can see that the model pretrained on our dataset and fine-tuned on the NYU-Depth V2 dataset achieves the best performance, both in error and accuracy. Figure 9 shows qualitative results. This demonstrates the usefulness of our dataset in improving algorithm performance in scene understanding tasks.

### 5.3 Benchmark and Diagnosis

In this section, we show benchmark results and provide a diagnosis of various common computer vision tasks using our synthetic dataset.

*Depth Estimation:* In the presented benchmark, we evaluated three state-of-the-art single-image depth estimation algorithms due to Eigen *et al.* [28, 29] and Liu *et al.* [71]. We evaluated those three algorithms with data generated from different settings including illumination intensities, colors, and object material properties. Table 4 shows a quantitative comparison. We see that both [29] and [28] are very sensitive to illumination conditions, whereas [71] is robust to illumination intensity, but sensitive to illumination color. All three algorithms are robust to different object materials. The reason may be that material changes do not alter the continuity of the surfaces. Note that [71] exhibits nearly the same performance on both our dataset and the NYU-Depth V2 dataset, supporting the assertion that our synthetic scenes are suitable for algorithm evaluation and diagnosis.

*Normal Estimation:* Next, we evaluated two surface normal estimation algorithms due to Eigen *et al.* [28] and Bansal *et al.* [3]. Table 5 summarizes our quantitative results. Compared to depth estimation, the surface normal estimation algorithms are stable to different illumination conditions as well as to different material properties. As in depth estimation, these two algorithms achieve comparable results on both our dataset and the NYU dataset.

*Semantic Segmentation:* Semantic segmentation has become one of the most popular tasks in scene understanding since the development and success of fully convolutional networks (FCNs). Given a single RGB image, the algorithm outputs a semantic label for every image pixel. We applied the semantic segmentation model described by Eigen *et al.* [28]. Since we have 129 classes of indoor objects whereas the model only has a maximum of 40 classes, we rearranged and reduced the number of classes to fit the prediction of the model. The algorithm achieves 60.5% pixel accuracy and 50.4 mIoU on our dataset.

*3D Reconstructions and SLAM:* We can evaluate 3D reconstruction and SLAM algorithms using images rendered from a sequence of camera views. We generated different sets of images on diverse synthesized scenes with various camera motion paths and backgrounds to evaluate the effectiveness of the open-source SLAM algorithm ElasticFusion [132]. A qualitative result is shown in Figure 10. Some scenes can be robustly reconstructed when we rotate the camera evenly and smoothly, as well as when both the background and foreground objects have rich textures. However, other reconstructed 3D meshes are badly fragmented due to the failure to register the current frame with previous frames due to fast moving cameras or the lack of textures. Our experiments indicate that our synthetic scenes with configurable attributes and background can be utilized to diagnose the SLAM algorithm, since we have full control of both the scenes and the camera trajectories.

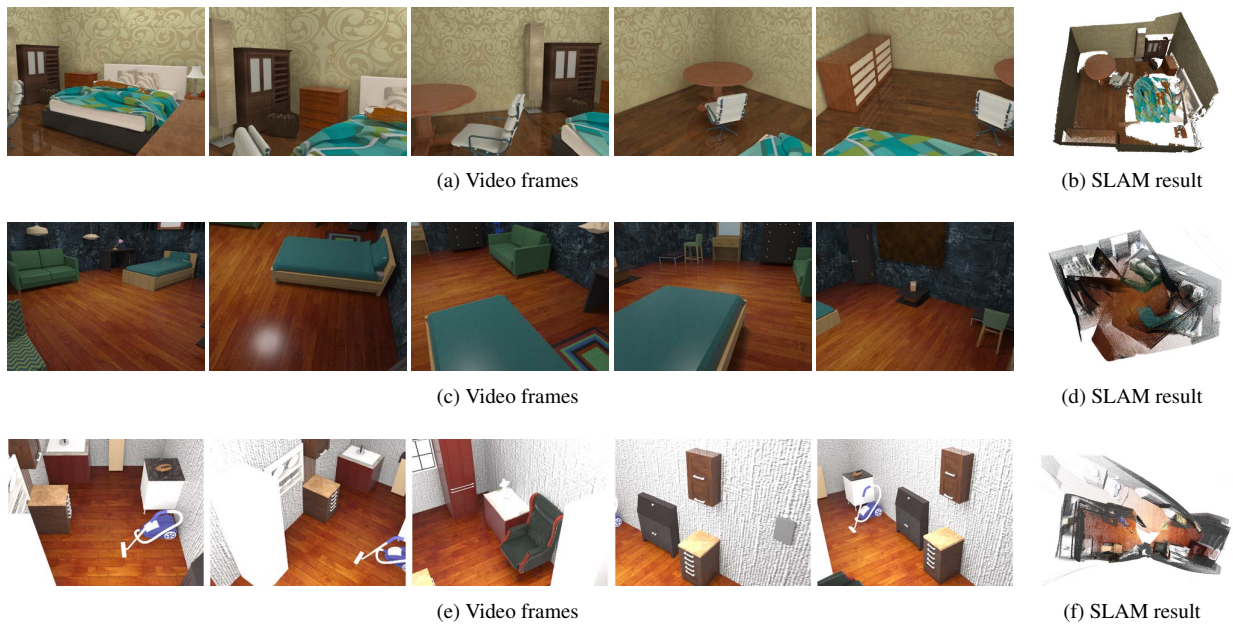

Fig. 10: Specifying camera trajectories, we can render scene fly-throughs as sequences of video frames, which may be used to evaluate SLAM reconstruction [132] results; *e.g.*, (a) (b) a successful reconstruction case and two failure cases due to (c) (d) a fast moving camera and (e) (f) untextured surfaces.

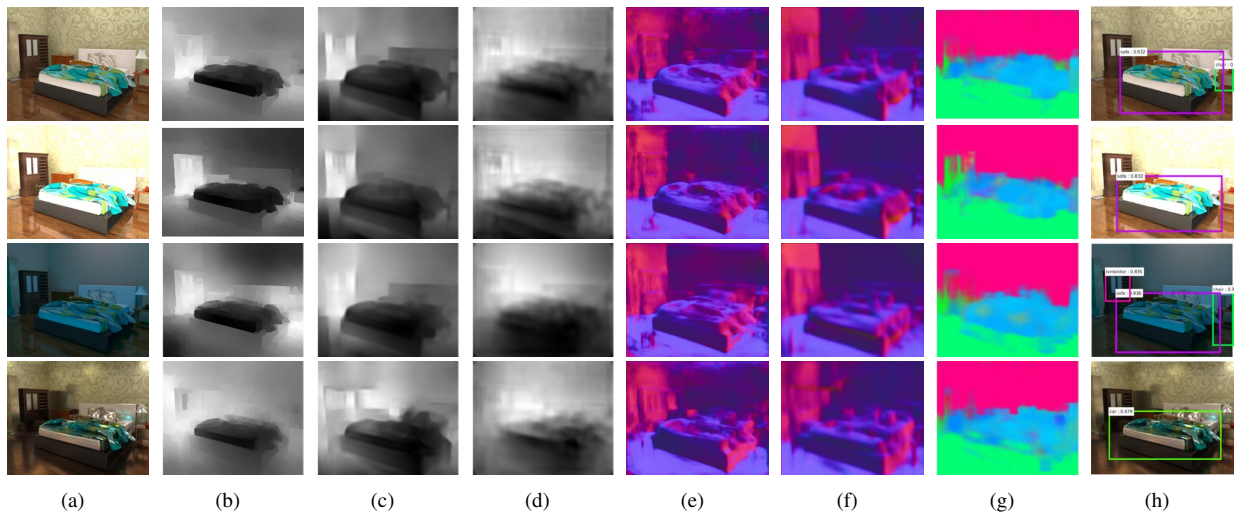

Fig. 11: Benchmark results. (a) Given a set of generated RGB images rendered with different illuminations and object material properties—(from top) original settings, high illumination, blue illumination, metallic material properties—we evaluate (b)–(d) three depth prediction algorithms, (e)–(f) two surface normal estimation algorithms, (g) a semantic segmentation algorithm, and (h) an object detection algorithm.

Table 4: Depth estimation. Intensity, color, and material represent the scene with different illumination intensities, colors, and object material properties, respectively.

| Setting | Method | Error | | | | | Accuracy | | |
|---------|--------|-------|-------|-------|---------------|-------------|----------------|------------------|------------------|
| | | Abs Rel | Sqr Rel | Log10 | RMSE (linear) | RMSE (log) | $\delta < 1.25$ | $\delta < 1.25^2$ | $\delta < 1.25^3$ |
| Original | [71] | 0.225 | 0.146 | 0.089 | 0.585 | 0.117 | 0.642 | 0.914 | 0.987 |
| | [29] | 0.373 | 0.358 | 0.147 | 0.802 | 0.191 | 0.367 | 0.745 | 0.924 |
| | [28] | 0.366 | 0.347 | 0.171 | 0.910 | 0.206 | 0.287 | 0.617 | 0.863 |
| Intensity | [71] | 0.216 | 0.165 | 0.085 | 0.561 | 0.118 | 0.683 | 0.915 | 0.971 |
| | [29] | 0.483 | 0.511 | 0.183 | 0.930 | 0.24 | 0.205 | 0.551 | 0.802 |
| | [28] | 0.457 | 0.469 | 0.201 | 1.01 | 0.217 | 0.284 | 0.607 | 0.851 |
| Color | [71] | 0.332 | 0.304 | 0.113 | 0.643 | 0.166 | 0.582 | 0.852 | 0.928 |
| | [29] | 0.509 | 0.540 | 0.190 | 0.923 | 0.239 | 0.263 | 0.592 | 0.851 |
| | [28] | 0.491 | 0.508 | 0.203 | 0.961 | 0.247 | 0.241 | 0.531 | 0.806 |
| Material | [71] | 0.192 | 0.130 | 0.08 | 0.534 | 0.106 | 0.693 | 0.930 | 0.985 |
| | [29] | 0.395 | 0.389 | 0.155 | 0.823 | 0.199 | 0.345 | 0.709 | 0.908 |
| | [28] | 0.393 | 0.395 | 0.169 | 0.882 | 0.209 | 0.291 | 0.631 | 0.889 |

Table 5: Surface Normal Estimation. Intensity, color, and material represent the setting with different illumination intensities, illumination colors, and object material properties, respectively.

| Setting | Method | Error | | | Accuracy | | |
|---------|--------|-------|--------|-------|----------|-------|-------|
| | | Mean | Median | RMSE | 11.25° | 22.5° | 30° |
| Original | [28] | 22.74 | 13.82 | 32.48 | 43.34 | 67.64 | 75.51 |
| | [3] | 24.45 | 16.49 | 33.07 | 35.18 | 61.69 | 70.85 |
| Intensity | [28] | 24.15 | 14.92 | 33.53 | 39.23 | 66.04 | 73.86 |
| | [3] | 24.20 | 16.70 | 32.29 | 32.00 | 62.56 | 72.22 |
| Color | [28] | 26.53 | 17.18 | 36.36 | 34.20 | 60.33 | 70.46 |
| | [3] | 27.11 | 18.65 | 35.67 | 28.19 | 58.23 | 68.31 |
| Material | [28] | 22.86 | 15.33 | 32.62 | 36.99 | 65.21 | 73.31 |
| | [3] | 24.15 | 16.76 | 32.24 | 33.52 | 62.50 | 72.17 |

*Object Detection:* The performance of object detection algorithms has greatly improved in recent years with the appearance and development of region-based convolutional neural networks. We apply the Faster R-CNN Model [102] to detect objects. We again need to rearrange and reduce the number of classes for evaluation. Figure 11 summarizes our qualitative results with a bedroom scene. Note that a change of material can adversely affect the output of the model—when the material of objects is changed to metal, the bed is detected as a "car".

## 6 Discussion

We now discuss in greater depth four topics related to the presented work.

*Configurable scene synthesis:* The most significant distinction between our work and prior work reported in the literature is our ability to generate large-scale *configurable* 3D scenes. But why is configurable generation desirable, given the fact that SUNCG [114] already provided a large dataset of manually created 3D scenes?

A direct and obvious benefit is the potential to generate *unlimited* training data. As shown in a recent report by Sun *et al.* [121], after introducing a dataset 300 times the size of ImageNet [25], the performance of supervised learning appears to continue to increase linearly in proportion to the increased volume of labeled data. Such results indicate the usefulness of labeled datasets on a scale even larger than SUNCG. Although the SUNCG dataset is large by today's standards, it is still a dataset limited by the need to manually specify scene layouts.

A benefit of using configurable scene synthesis is to diagnose AI systems. Some preliminary results were reported in this paper. In the future, we hope such methods can assist in building explainable AI. For instance, in the field of causal reasoning [90], causal induction usually requires turning on and off specific conditions in order to draw a conclusion regarding whether or not a causal relation exists. Generating a scene in a controllable manner can provide a useful tool for studying these problems.

Furthermore, a configurable pipeline may be used to generate various virtual environment in a controllable manner in order to train virtual agents situated in virtual environments to learn task planning [69, 155] and control policies [53, 130].

*The importance of the different energy terms:* In our experiments, the learned weights of the different energy terms indicate the importance of the terms. Based on the ranking from the largest weight to the smallest, the energy terms are 1) distances between furniture pieces and the nearest wall, 2) relative orientations of furniture pieces and the nearest wall, 3) supporting relations, 4) functional group relations, and 5) occlusions of the accessible space of furniture by other furniture. We can regard such rankings learned from training data as human preferences of various factors in indoor layout designs, which is important for sampling and generating realistic scenes. For example, one can imagine that it is more important to have a desk aligned with a wall (relative distance and orientation), than it is to have a chair close to a desk (functional group relations).

*Balancing rendering time and quality:* The advantage of physically accurate representation of colors, reflections, and shadows comes at the cost of computation. High quality rendering (*e.g.*, rendering for movies) requires tremendous amounts of CPU time and computer memory that is practical only with distributed rendering farms. Low quality settings are prone to granular rendering noise due to stochastic sampling. Our comparisons between rendering time and rendering quality serve as a basic guideline for choosing the values of the rendering parameters. In practice, depending on the complexity of the scene (such as the number of light sources and reflective objects), manual adjustment is often needed in large-scale rendering (*e.g.*, an overview of a city) in order to achieve the best trade-off between rendering time and quality. Switching to GPU-based ray tracing engines is a promising alternative. This direction is especially useful for scenes with a modest number of polygons and textures that can fit into a modern GPU memory.

*The speed of the sampling process:* Using our computing hardware, it takes roughly 3–5 minutes to render a $640 \times 480$-pixel image, depending on settings related to illumination, environments, and the size of the scene. By comparison, the sampling process consumes approximately 3 minutes with the current setup. Although the convergence speed of the Monte Carlo Markov chain is fast enough relative to photorealistic rendering, it is still desirable to accelerate the sampling process. In practice, to speed up the sampling and improve the synthesis quality, we split the sampling process into five stages: (i) Sample the objects on the wall, *e.g.*, windows, switches, paints, and lights, (ii) sample the core functional objects in *functional groups* (*e.g.*, desks and beds),

(iii) sample the objects that are associated with the core functional objects (*e.g.*, chairs and nightstands), (iv) sample the objects that are not paired with other objects (*e.g.*, wardrobes and bookshelves), and (v) Sample small objects that are supported by furniture (*e.g.*, laptops and books). By splitting the sampling process in accordance with functional groups, we effectively reduce the computational complexity, and different types of objects quickly converge to their final positions.

## 7 Conclusion and Future Work

Our novel learning-based pipeline for generating and rendering configurable room layouts can synthesize unlimited quantities of images with detailed, per-pixel ground truth information for supervised training. We believe that the ability to generate room layouts in a controllable manner can benefit various computer vision areas, including but not limited to depth estimation [28, 29, 66, 71], surface normal estimation [3, 28, 128], semantic segmentation [17, 74, 87], reasoning about object-supporting relations [34, 68, 112, 148], material recognition [7–9, 134], recovery of illumination conditions [5, 48, 63, 73, 86, 88, 89, 108, 145], inference of room layout and scene parsing [19, 24, 43, 52, 57, 67, 78, 135, 147], determination of object functionality and affordance [4, 40, 44, 54, 59, 61, 62, 85, 107, 116, 142, 147, 153], and physical reasoning [133, 134, 148, 149, 154, 156]. In additional, we believe that research on 3D reconstruction in robotics and on the psychophysics of human perception can also benefit from our work.

Our current approach has several limitations that we plan to address in future research. First, the scene generation process can be improved using a multi-stage sampling process; *i.e.*, sampling large furniture objects first and smaller objects later, which can potentially improve the scene layout. Second, we will consider modeling human activity inside the generated scenes, especially with regard to functionality and affordance. Third, we will consider the introduction of moving virtual humans into the scenes, which can provide additional ground truth for human pose recognition, human tracking, and other human-related tasks. To model dynamic interactions, a Spatio-Temporal AOG (ST-AOG) representation is needed to extend the current spatial representation into the temporal domain. Such an extension would unlock the potential to further synthesize outdoor environments, although a large-scale, structured training dataset would be needed for learning-based approaches. Finally, domain adaptation has been shown to be important in learning from synthetic data [76, 106, 123]; hence, we plan to apply domain adaptation techniques to our synthetic dataset.